\documentclass[10pt,twocolumn,letterpaper]{article}

\usepackage{cvpr}
\usepackage{times}
\usepackage{epsfig}
\usepackage{graphicx}
\usepackage{amsmath}
\usepackage{amssymb}
\usepackage{caption}
\usepackage{algorithm}
\usepackage{algorithmic}
\usepackage{multirow}
\usepackage{subcaption}
\usepackage{dsfont}

\usepackage[pagebackref=true,breaklinks=true,letterpaper=true,colorlinks,bookmarks=false]{hyperref}
\usepackage{relsize}
\cvprfinalcopy 

\def\cvprPaperID{394} 

\ifcvprfinal\fi
\begin{document}
\title{Hand Keypoint Detection in Single Images using Multiview Bootstrapping}
\author{Tomas Simon \hspace{0.25in} Hanbyul Joo \hspace{0.25in} Iain Matthews \hspace{0.25in} Yaser Sheikh\vspace{0.05in}\\
Carnegie Mellon University\\
{\tt\small \{tsimon,hanbyulj,iainm,yaser\}@cs.cmu.edu}}
\maketitle
\thispagestyle{empty}

\begin{abstract}
We present an approach that uses a multi-camera system to train fine-grained detectors for keypoints that are prone to occlusion, such as the joints of a hand. We call this procedure {\em multiview bootstrapping:} first, an initial keypoint detector is used to produce noisy labels in multiple views of the hand. The noisy detections are then triangulated in 3D using multiview geometry or marked as outliers. Finally, the reprojected triangulations are used as new labeled training data to improve the detector. We repeat this process, generating more labeled data in each iteration. We derive a result analytically relating the minimum number of views to achieve target true and false positive rates for a given detector. The method is used to train a hand keypoint detector for single images. The resulting keypoint detector runs in realtime on RGB images and has accuracy comparable to methods that use depth sensors. The single view detector, triangulated over multiple views, enables 3D markerless hand motion capture with complex object interactions.
\end{abstract}
\vspace{-24pt}

\section{Introduction}
While many approaches to image-based face and body keypoint localization exist, there are no markerless hand keypoint detectors that work on RGB images in the wild. This is surprising given the important role that hands play in our daily activities---they are the way we interact with the world: we use tools, we play instruments, we touch, and we gesture. A method that can localize hand joints in RGB images (without requiring depth) would enable new analyses of human motion on the largest existing sources of visual data (e.g., YouTube and Netflix), as well as new applications in HCI and robotics. We present a method that enables realtime 2D hand tracking in single view video and 3D hand motion capture, as shown in Fig.~\ref{fig:teaser}.

\begin{figure}[t]
\includegraphics[width=\linewidth]{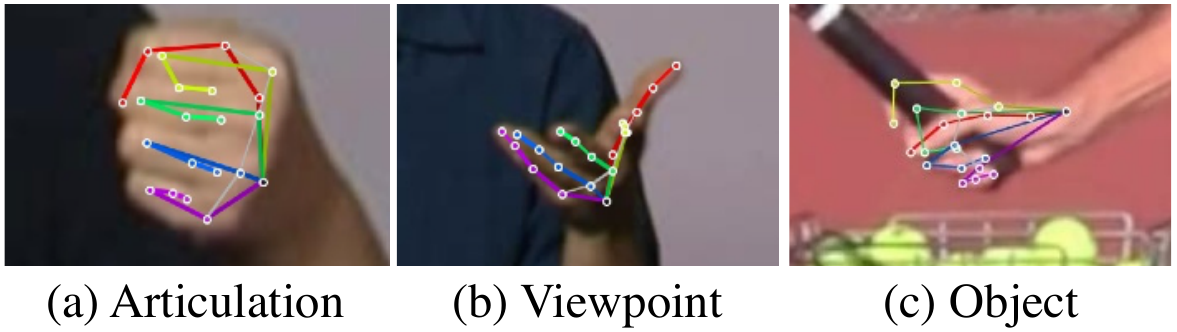}
\caption{Hand annotation is difficult in single images because joints are often occluded due to (a) articulations of other parts of the hand, (b) a particular viewing angle, or (c) objects that the hand is grasping.}
\label{fig:training_sets}
\end{figure}

Unlike the face and body, large datasets of annotated keypoints do not exist for hands. Generating such datasets presents a major challenge compared to the face or body. Due to heavy occlusions, even {\em manual} keypoint annotations are difficult to get right: for the keypoints that are occluded, the annotated locations are at best an educated guess. Fig.~\ref{fig:training_sets} shows examples of manually annotated images that contain self-occlusion due to articulation, self-occlusion due to viewpoint, and occlusion by a grasped object. In each case, several keypoints had to be estimated by the annotator, increasing annotation time and cost while reducing accuracy.

In this paper, we present an approach to boost the performance of a given keypoint detector using a multi-camera setup. This approach, which we refer to as multiview bootstrapping, is based on the following observation: even if a particular image of the hand has significant occlusion, there often exists an unoccluded view. Multiview bootstrapping systematizes this insight to produce a more powerful hand detector, that we demonstrate generalizes beyond the capture setup. In particular, it allows a weak detector, trained on a small annotated dataset, to localize subsets of keypoints in \emph{good} views and uses robust 3D triangulation to filter out incorrect detections. Images where severe occlusions exists are then labeled by reprojecting the triangulated 3D hand joints. By including these newly generated annotations in the training set, we iteratively improve the detector, obtaining more and more accurate detections at each iteration. This approach generates geometrically consistent hand keypoint annotations using multiview constraints as an external source of supervision. In this way, we can label images that are difficult or impossible to annotate due to occlusion. We demonstrate that multiview bootstrapping produces hand keypoint detectors for RGB images that rival the performance of RGB-D hand keypoint detectors. We further show that applying this single view detector in a multi-camera setup allows markerless 3D hand reconstructions in unprecedented scenarios, including challenging manipulation of objects, musical performances, and multiple interacting people.

\section{Related Work}

\begin{figure*}[t]
\centering
    \includegraphics[width=0.98\linewidth,clip=true,trim=-10 16 34 5pt]{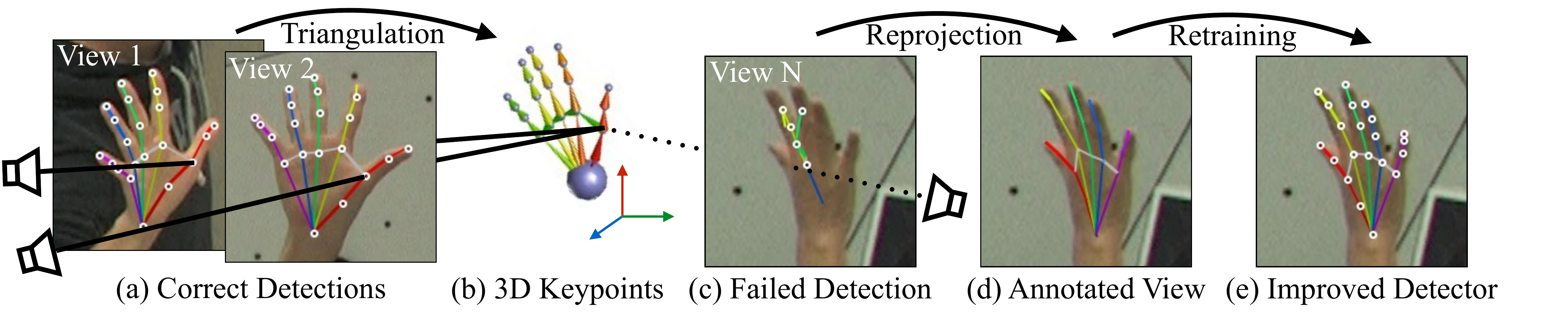}
  \caption{ Multiview Bootstrapping. (a) A multiview system provides views of the hand where keypoint detection is easy, which are used to triangulate (b) the 3D position of the keypoints. Difficult views with (c)  failed detections can be (d) annotated using the reprojected 3D keypoints, and used to retrain (e) an improved detector that now works on difficult views. \label{fig:hand_overview}}
\end{figure*}
Early work in hand pose estimation originally considered RGB data, with Rehg and Kanade~\cite{Rehg-94} exploring vision-based Human-Computer Interaction (HCI) applications. Most methods were brittle, based on fitting complex 3D models with strong priors, including e.g., physics or dynamics~\cite{Lu-03}, multiple hypotheses~\cite{Stenger-06}, or analysis-by-synthesis~\cite{LaGorce-11}. Cues such as silhouettes, edges, skin color, and shading were demonstrated in controlled environments with restricted poses and simple motions. The method of Wang and Popovi\'c~\cite{Wang-09} lifted some of these restrictions, but required a specialized colored glove. Multiview RGB methods are often similarly based on fitting sophisticated mesh models~(e.g., \cite{Ballan-12, Sridhar-13}) and show excellent accuracy, but again under highly controlled conditions. 

With the introduction of commodity depth sensors, single-view depth-based hand pose estimation became the major focus of research, resulting in a large number of depth-based methods. These are broadly classifiable into generative methods~\cite{Oikonomidis-12}, discriminative methods~\cite{Tang-14,Tompson2014b, Keskin-12,Xu-13,Sun-15,Wan-16}, or hybrid methods~\cite{Sridhar-13, Sharp-15, Sridha-15, Tzionas-16, Ye-16}. Recently, the hybrid method of Sharp et al.~\cite{Sharp-15} demonstrated practical performance across a large range, but there are still difficult cases such as hand-hand interactions and hand-object interactions. Discriminative and hybrid approaches to depth-based hand pose estimation rely heavily on synthetic data~\cite{Supancic-15}. Oberwerger et al.~\cite{Oberweger-15a} use feedback loops to generate synthetic training data for hand pose estimation, motivated by the same principles as our work, but focus on generating depth images. The semi-automatic data annotation scheme presented in~\cite{Oberweger-16} is also similar in motivation, however, our approach uses multiview geometry and keypoint detection to provide automated supervision. 

Discriminative methods, especially approaches that rely on deep architectures, require large annotated training sets. These datasets are relatively easy to synthesize for depthmaps, but present significant challenges for RGB as rendering is far more complicated, requiring photorealistic appearance and lighting. Multiview bootstrapping is an approach that allows the generation of large annotated datasets using a weak initial detector. This, in turn, enables the create of the first realtime hand keypoint detector for RGB images ``in-the-wild".

\section{Multiview Bootstrapped Training}

A keypoint detector $d(\cdot)$ maps a cropped input image patch $\mathbf{I} \,{\in}\, \mathds{R}^{w\times h\times 3}$ to $P$ keypoint locations $\mathbf{x}_p\,{\in}\,\mathds{R}^2$, each with an associated detection confidence $c_p$: 
\begin{equation}
d(\mathbf{I}) \mapsto \left\{ (\mathbf{x}_p, c_p) \ \textrm{for}\ p \in [1\dotsc P]\right\}.
\end{equation}
Each point $p$ corresponds to a different landmark (e.g.,~the tip of the thumb, the tip of the index finger, see Fig.~\ref{fig:confmaps}a), and we assume that only a single instance of the object is visible in $\mathbf{I}$. The detector is trained on images with corresponding keypoint annotations, $\left(\mathbf{I}^f, \{ \mathbf{y}^f_p \}\right)$, where $f$ denotes a particular image frame, and the set $\{ \mathbf{y}^f_p \in \mathds{R}^2\}$ includes all labeled keypoints for the image $\mathbf{I}^f$. An initial training set $\mathcal{T}_0$ having $N_0$ training pairs,
\begin{equation}
\mathcal{T}_0 := \left\{ \left(\mathbf{I}^f, \{ \mathbf{y}^f_p \}\right)\ \textrm{for}\ f \in[1\dotsc N_0]\right\},
\end{equation}
is used to train an initial detector $d_0$ with, e.g.,~stochastic gradient descent,
\begin{equation}
d_0 \leftarrow \operatorname{train}(\mathcal{T}_0).
\end{equation}
Given the initial keypoint detector $d_0$ and a dataset of unlabeled multiview images, our objective is to use the detector to generate a set of labeled images, $\mathcal{T}_1$, which can be used to train an {\em improved} detector, $d_1$, using all available data:
\begin{equation}
d_1 \leftarrow \operatorname{train}(\mathcal{T}_0 \cup \mathcal{T}_1).
\end{equation}
To improve upon the detector $d_0$, we need an external source of supervision to ensure $\mathcal{T}_1$ contains information not already present in $\mathcal{T}_0$. 
We propose to use verification via multiview geometry as this source. The key here is that detection is easier in some views than others:~if a point is successfully localized in at least two views, the triangulated 3D position can be reprojected onto other images, providing a new 2D annotation for the views on which detection failed. This process is shown in Fig.~\ref{fig:hand_overview}, where the detector succeeds on easy views (Fig.~\ref{fig:hand_overview}a) but fails on more challenging views (Fig.~\ref{fig:hand_overview}c). However, by triangulating the correctly detected viewpoints, we can generate training data particularly for those views in which the detector is currently failing.

The overall procedure for multiview bootstrapping is described in Algorithm~\ref{alg:mvbs}, where we denote by $\{\mathbf{I}_v^f : v\,{\in}\,[1\dotsc V], f\,{\in},[1\dotsc F]\}$ the input set of unlabeled multiview image frames, with $v$ iterating over the $V$ camera views, and $f$ iterating over $F$ distinct frames (i.e., time instants, so one frame represents $V$ images). There are three main parts to the process detailed in the following subsections: (1) for every frame, the algorithm first runs the current detector on every camera view independently (Fig.~\ref{fig:hand_overview}a,c) and robustly triangulates the point detections (Fig.~\ref{fig:hand_overview}b); (2) the set of frames is then sorted according to a score to select only correctly triangulated examples, and (3) the $N$-best frames are used to train a new detector by reprojecting the correctly triangulated points onto all views (Fig.~\ref{fig:hand_overview}d), producing approximately $V$ training images for each of the $N$ selected frames. The entire process can then be iterated with the newly trained detector (Fig.~\ref{fig:hand_overview}e). 
\begin{algorithm}[t]
\caption{Multiview Bootstrapping}
\label{alg:mvbs}
\begin{algorithmic} 
\STATE\textbf{Inputs:}\\ \textbullet~Unlabeled images: $\{ \mathbf{I}_v^f\ \textrm{for} \ v \,{\in}\, \textrm{views}, f \,{\in}\, \textrm{frames} \}$ \\\textbullet~Keypoint detector: $d_0(\mathbf{I}) \mapsto \{ (\mathbf{x}_p, c_p) \  \textrm{for} \ p \in \textrm{points}\}$  \\\textbullet~Labeled training data: $\mathcal{T}_{0}$
\STATE\textbf{for} iteration $i$ in $0$ to $K$:
\STATE\hspace{8pt}1. Triangulate keypoints from weak detections
\STATE\hspace{12pt} \textbf{for} every frame $f$:
\STATE\hspace{8pt}\hspace{8pt} \hspace{8pt} $(a)$ Run detector $d_i(\mathbf{I}_v^f)$ on all views $v$ (Eq.~\eqref{eq:rundetector})
\STATE\hspace{8pt}\hspace{8pt} \hspace{8pt} $(b)$ Robustly triangulate keypoints (Eq.~\eqref{eq:triangulate})
\STATE \hspace{8pt}2. Score and sort triangulated frames (Eq.~\eqref{eq:scoring_sort})
\STATE \hspace{8pt}3. Retrain with $N$-best reprojections (Eq.~\eqref{eq:hand_reprojecting})
\STATE \hspace{8pt}\hspace{8pt} $d_{i+1} \leftarrow \operatorname{train}(\mathcal{T}_0 \cup \mathcal{T}_{i+1}$)
\STATE\textbf{Outputs:} Improved detector $d_K(\cdot)$ and training set $\mathcal{T}_{K}$
\end{algorithmic}
\end{algorithm}

\subsection{Triangulating Keypoints from Weak Detections}
Given $V$ views of an object in a particular frame $f$, we run the current detector $d_i$ (trained on set $\mathcal{T}_i$) on each image $\mathbf{I}_v^f$, yielding a set $\mathcal{D}$ of 2D location candidates:
\begin{equation}\label{eq:rundetector}
\mathcal{D} \leftarrow \{ d_{i}( \mathbf{I}_v^f ) \textrm{ for } v \in [1\dotsc V]\}.
\end{equation}
For each keypoint $p$, we have $V$ detections $(\mathbf{x}_p^v, c_p^v)$, where $\mathbf{x}_p^v$ is the detected location of point $p$ in view $v$ and $c_p^v\,{\in}\,[0,1]$ is a confidence measure (we omit the frame index for clarity). To robustly triangulate each point $p$ into a 3D location, we use RANSAC~\cite{Fischler-81} on points in $\mathcal{D}$ with confidence above a detection threshold $\lambda$. Additionally, we use a $\sigma{=}4$ pixel reprojection error to accept RANSAC inliers. With this set of inlier views for point $p$, we minimize~\cite{ceres-solver} the reprojection error to obtain the final triangulated position,
\begin{equation}\label{eq:triangulate}
\mathbf{X}_p^f = \operatorname{arg} \underset{\mathbf{X}}{\operatorname{min}} \sum_{v{\in}\mathcal{I}_p^f} || \mathcal{P}_v(\mathbf{X}) - \mathbf{x}_p^v ||_2^2,
\end{equation}
where $\mathcal{I}_p^f$ is the inlier set, with $\mathbf{X}^f_p\in\mathds{R}^3$ the 3D triangulated keypoint $p$ in frame $f$, and $\mathcal{P}_v( \mathbf{X})\in\mathds{R}^2$ denotes projection of 3D point $\mathbf{X}$ into view $v$.
Given calibrated cameras, this 3D point can be reprojected into any view (e.g., those in which the detector failed) and serve as a new training label.

To improve robustness specifically for hands, we reconstruct entire fingers simultaneously. We triangulate all landmarks of each finger (4 points) at a time, and use the average reprojection error of all 4 points to determine RANSAC inliers. This procedure is more robust because errors in finger detections are correlated: e.g.,~if the knuckle is incorrectly localized, then dependent joints in the kinematic chain---the inter-phalangeal joints and finger tip---are unlikely to be correct. This reduces the number of triangulated keypoints (because the entire finger needs to be correct in the same view) but it further reduces the number of false positives, which is more important so that we do not train with incorrect labels.

\subsection{Scoring and Sorting Triangulated Frames}
It is crucial that we do not include erroneously labeled frames as training data, especially if we iterate the procedure, as subsequent iterations will fail in a {\em geometrically consistent} way across views---a failure which cannot be detected using multi-view constraints. We therefore conservatively pick a small number of reliable triangulations.

Our input is video and consecutive frames are therefore highly correlated. Instead of uniform temporal subsampling, we pick the ``best'' frame for every window of $W$ frames (e.g., $W{=}15$ or $W{=}30$), defining the ``best'' as that frame with maximum sum of detection confidences for the inliers, i.e., \vspace{-4pt}
\begin{equation}
\operatorname{score}( \{ \mathbf{X}_p^f\} ) = \sum_{p{\in}[1\dotsc P]} \sum_{v {\in} \mathcal{I}^f_p} c_p^v.
\label{eq:scoring_sort}
\end{equation}\vspace{-2pt}
We sort all the remaining frames in descending order according to their $\operatorname{score}$, to obtain an ordered sequence of frames, $[s_1, s_2, \dotsc s_{F'}]$, where $F'$ is the number of subsampled frames and $s_i$ is the ordered frame index.

We manually verify that there are no obvious errors in the frames to be used as training data, and train a new detector. While visual inspection of the training set may seem onerous, in our experience this is the least time consuming part of the training process. It typically takes us one or two minutes to inspect the top $100$ frames. Crowdsourcing such a verification step for continuous label generation is an interesting future direction, as verification is easier than annotation. In practice, we find that this manual effort can be almost eliminated by automatically removing questionable triangulations using a number of heuristics: (1) average number of inliers, (2) average detection confidence, (3) difference of per-point velocity with median velocity between two video frames, (4) anthropomorphic limits on joint lengths\footnote{We use thresholds larger than the maximum bone lengths given in the survey of Greiner~\cite{Greiner-91}, specifically 15~cm for the metacarpals, 9~cm for the proximal phalanges, and 5~cm for the remaining bones.}, and (5) complete occlusion, as determined by camera ray intersection with body joints. Additionally, we require at least 3 inliers for any point to be valid. 

\subsection{Retraining with $N$-best Reprojections}
We use the $N$-best frames according to this order to define a new set of training image-keypoint pairs for the next iteration $i{+}1$ detector,
\begin{align}
\mathcal{T}_{i+1} = \left\{ \left(\mathbf{I}_v^{s_n}, \{\mathcal{P}_v( \mathbf{X}_p^{s_n}) : v{\in}[1\dotsc V],\,p{\in} [1\dotsc P] \} \right)\right. \nonumber \\\left.\textrm{ for } n{\in}[1\dotsc N]\right\},
\label{eq:hand_reprojecting}
\end{align}
where $\mathcal{P}_v( \mathbf{X}_p^{s_n} )$ denotes projection of point $p$ for frame index $s_n$ into view $v$, and we aim for about $N{=}100$ frames for every $3$ minutes of video. Note that $100$ frames yields roughly $100\cdot \frac{V}{2}\approxeq 1500$ training samples, one for each unoccluded viewpoint. Finally, we train a new detector using the expanded training set as $d_{i+1}{\leftarrow}\operatorname{train}(\mathcal{T}_{0}\cup\mathcal{T}_{i+1})$.
\begin{figure}
\includegraphics[width=\linewidth,clip=true,trim=16pt 10pt 0pt 4pt]{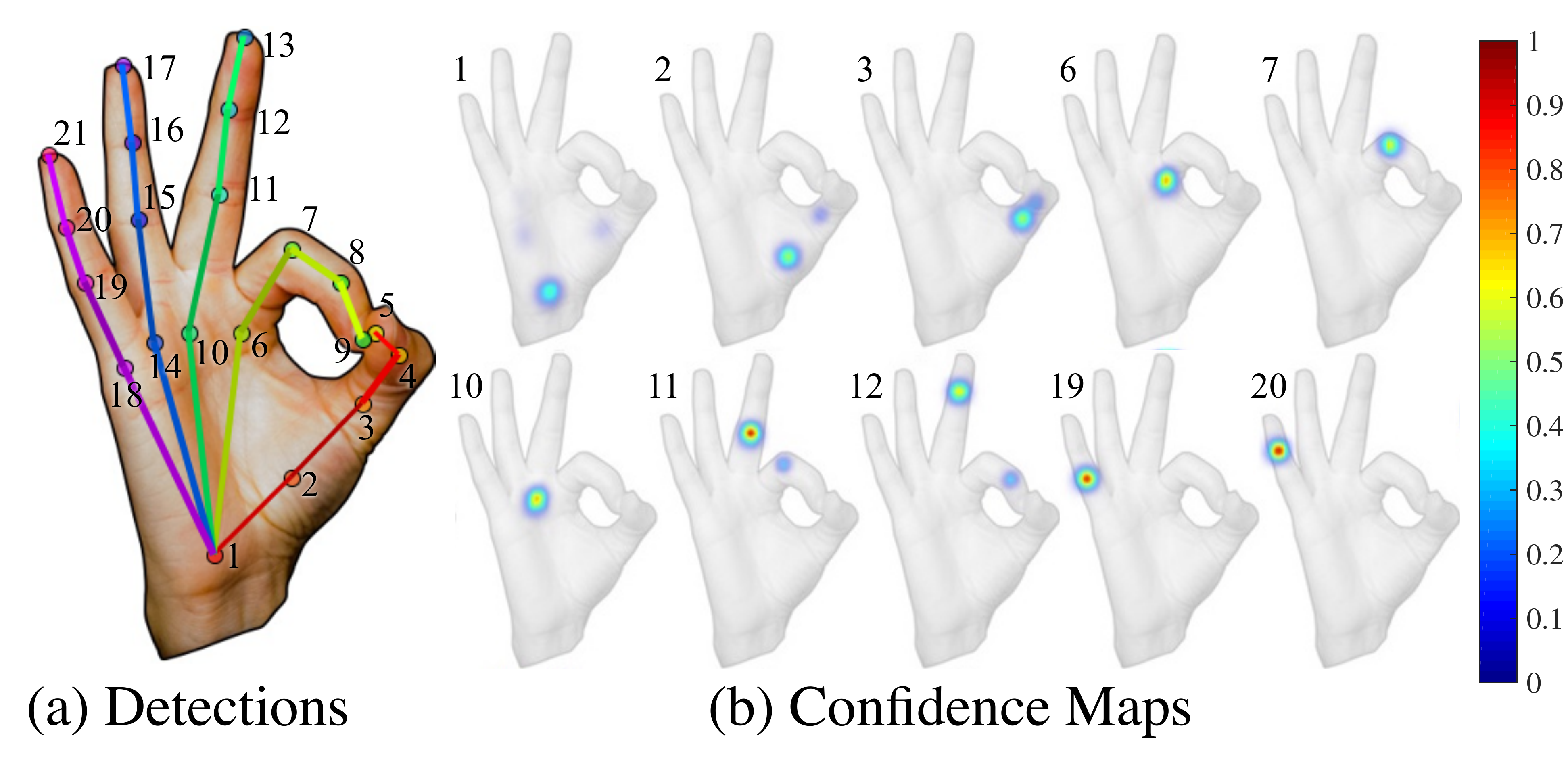}
\caption{(a)~Input image with 21 detected keypoints. (b)~Selected confidence maps produced by our detector, visualized as a ``jet'' colormap overlaid on the input.}
\label{fig:confmaps}
\end{figure}

\section{Detection Architecture}
For the detectors $d_i$, we follow the architecture of Convolutional Pose Machines (CPMs)~\cite{Wei2016}, with some modification. CPMs predict a confidence map for each keypoint, representing the keypoint's location as a Gaussian centered at the true position. The predicted confidence map corresponds to the size of the input image patch, and the final position for each keypoint is obtained by finding the maximum peak in each confidence map (see Fig.~\ref{fig:confmaps}b). \vspace{5pt}

\noindent \textbf{Keypoint Detection via Confidence Maps.} In contrast to~\cite{Wei2016}, we use the convolutional stages of a pre-initialized VGG-19 network~\cite{Simonyan-14} up to $\operatorname{conv4\_4}$ as a feature extractor, with two additional convolutions producing 128-channel features~$\mathbf{F}$. For an input image patch of size $w{\times}h$, the resulting size of the feature map $\mathbf{F}$ is $w'{\times}h'\times128$, with $w'{=}\frac{w}{8}$ and $h'{=}\frac{h}{8}$. There are no additional pooling or downsampling stages, so the final stride of the network is also $8$. This feature map extraction is followed by a prediction stage that produces a set of $P$ confidence or {\em score} maps, $\mathbf{S}^1{=}\{\mathbf{S}^1_1\dots \mathbf{S}^1_P\}$, one score map $\mathbf{S}^1_p\in \mathds{R}^{w'{\times}h'}$ for each keypoint $p$. Each stage after the first takes as input the score maps from the previous stage, $\mathbf{S}^{t-1}$, concatenated with the image features~$\mathbf{F}$, and produces $P$ new score maps~$\mathbf{S}^{t}$, one for each keypoint. We use $6$ sequential prediction stages, taking the output at the final stage, $\mathbf{S}^6$. We resize these maps to the original patch size ($w\times h$) using bicubic resampling, and extract each keypoint location as the pixel with maximum confidence in its respective map.
We also modify the loss function in~\cite{Wei2016} to be a weighted $L_2$ loss to handle missing data, where the weights are set to zero if annotations for a keypoint are missing (e.g., if triangulation for that point fails).

\label{sect:architecture}

\vspace{5pt}
\noindent \textbf{Hand Bounding Box Detection.} Our keypoint detector assumes that the input image patch $\mathbf{I}{\in}\mathds{R}^{w\times h \times 3}$ is a crop around the right hand. This is an important detail: to use the keypoint detector in any practical situation, we need a way to generate this bounding box. We directly use the body pose estimation models from~\cite{Wei2016} and~\cite{Zhe-17}, and use the wrist and elbow position to approximate the hand location, assuming the hand extends $0.15$ times the length of the forearm in the same direction. During training, we crop a square patch of size $2.2B$, where $B$ is the maximum dimension of the tightest bounding box enclosing all hand joints (see Fig.~\ref{fig:mpi_nzsl_qualitative} for example crops of this size). At test time, we approximate $B\,{=}\,0.7H$ where $H$ is the length of the head ``joint'' (top of head to bottom). This square patch is resized to $w\,{=}\,368$ and $h\,{=}\,368$, which serves as input to the network. To process left hands, we flip the image left-to-right and treat it as a right hand.
\begin{figure}[t]
\centering
\includegraphics[width=0.96\linewidth,clip=true,trim=0pt 10pt 0pt 36pt]{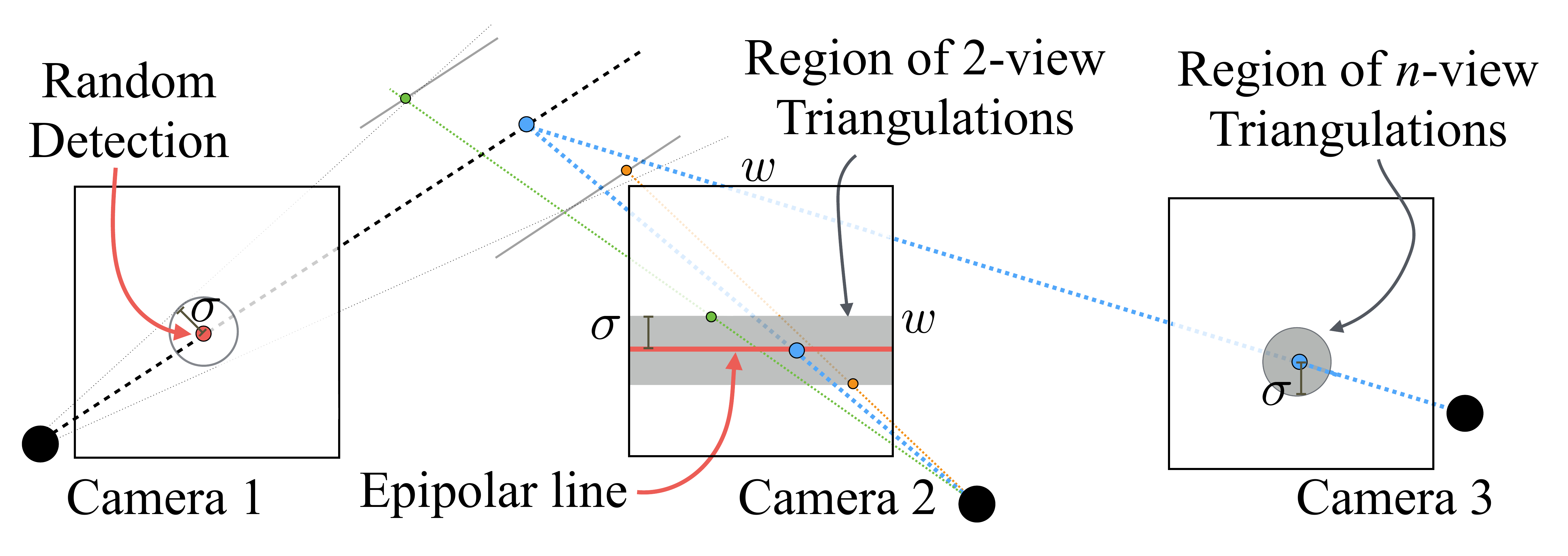}
\caption{Approximate area within which a random detection will be successfully triangulated for 2 and 3+ views with an inlier threshold of $\sigma$ pixels.}
\label{fig:epipolar}
\end{figure}
\section{When does Multiview Bootstrapping Work?}

In this section, we derive results that allow us to determine how many camera views are necessary for multiview bootstrapping to work for a given detector, or conversely how accurate an initial detector has to be for multiview bootstrapping to work for a given number of cameras. Detailed derivations and assumptions are included in the supplementary document.

Let us first define the quality of a detector $d_0$ as its {\em Probability of Correct Keypoint} or PCK
: the probability that a predicted keypoint is within a distance threshold~$\sigma$ of its true location. For a particular keypoint~$p$, we denote it by $\operatorname{PCK}^p_\sigma(d_0)$ and approximate it on a testing set $\mathcal{T}$~as
\begin{eqnarray}
\label{eq:pck}
\operatorname{PCK}^p_\sigma(d_0) := \frac{1}{|\mathcal{T}|}\sum_{\mathcal{T}} \delta\left( || \mathbf{x}^f_p - \mathbf{y}_p^f ||_2 < \sigma \right),
\end{eqnarray}
for $\mathbf{x}_p^f \,{\in}\, d_0(\mathbf{I}^f)$ the $p$-th keypoint prediction on image $\mathbf{I}^f$ and~$\mathbf{y}^f_p$ its true location, with $\delta(\cdot)$ the indicator function.
For multiview bootstrapping to succeed, we need a low false positive rate in accepting erroneous triangulations as valid. We derive three results that quantify the probability of erroneous triangulations (that do not correspond to correct detections) for varying detector quality and camera setups.

We first define some preliminary quantities. The probability $q_2$ of a spurious  triangulation within a distance $\sigma$ by two points sampled uniformly on an image square of size $w\times w$ is approximately $\frac{2\sigma w}{w^2}$ (assuming rectified stereo pairs as shown in Fig.~\ref{fig:epipolar}). The probability that this spurious triangulation is supported by a third view is bounded by $\frac{\pi \sigma^2}{w^2}$. Further, the probability that at least $n-2$ points in the remaining $V{-}2$ views support this spurious triangulation is $p_{n-2}=\operatorname{Pr}(X \geq n{-}2)$ where $X \sim \operatorname{B}(V{-}2, \pi \frac{\sigma^2}{w^2})$ is a binomial random variable\footnote{For $X\sim \operatorname{B}(N,p)$ binomial with parameters $N$ and $p$, then $\operatorname{Pr}(X{=}k) = \begin{pmatrix}N\\ k\end{pmatrix}p^k (1-p)^{N-k}$ and $\operatorname{Pr}(X{\geq}k)=\sum_{l=k}^N \operatorname{Pr}(X{=}l)$.} denoting the number of inliers in the remaining $V{-}2$ views for the spurious triangulation. Finally, $q_n \approxeq q_2 \cdot p_{n-2}$ is the probability that, for a given pair of views, you find $n$ inliers that support a spurious triangulation.\\

\begin{figure}[t]
\includegraphics[width=\linewidth]{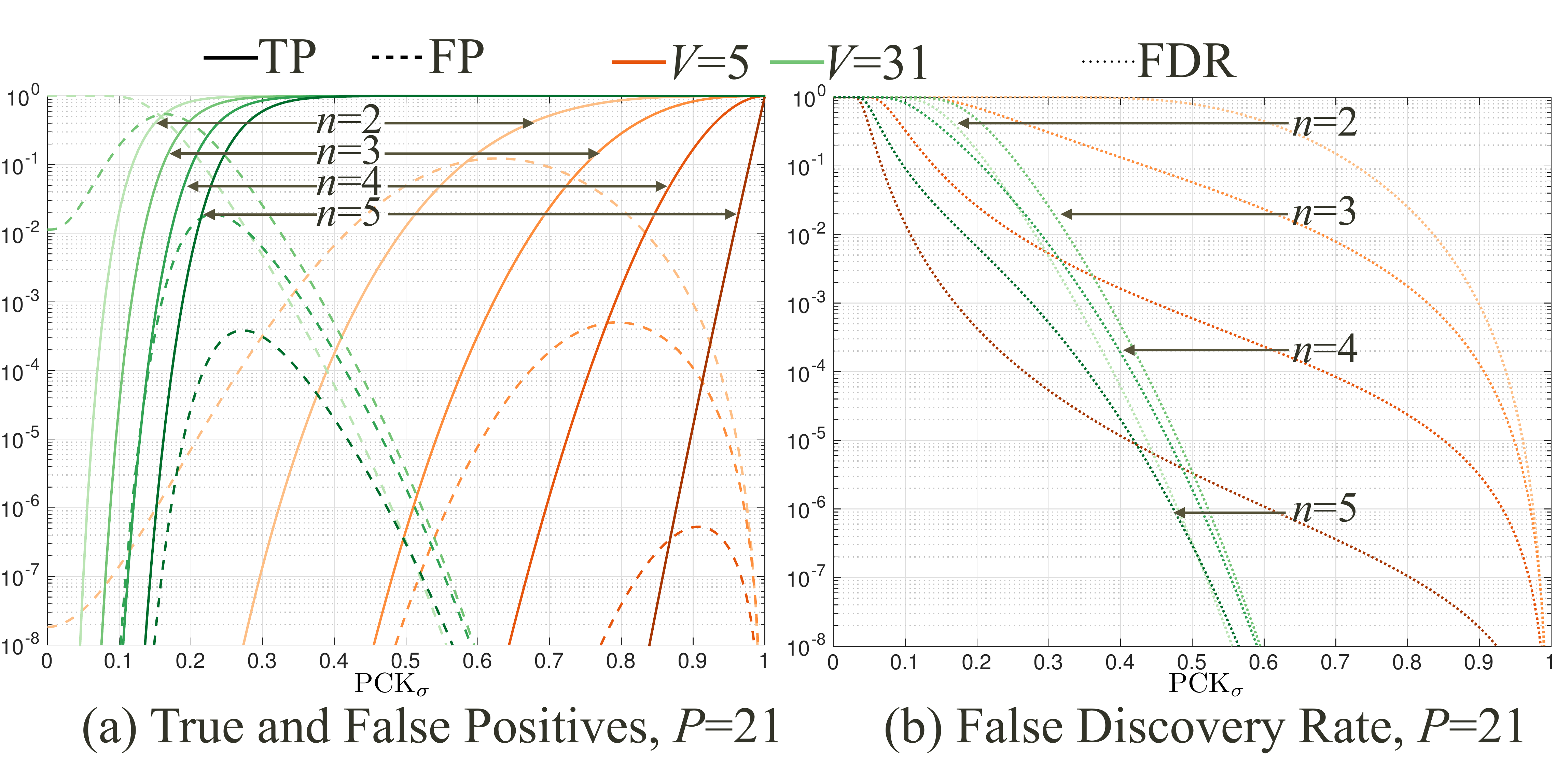}
\caption{(a) TP and FP for different values of $\operatorname{PCK}_\sigma$, and different number of $n$ inliers. A setup with 5 cameras ($V=5$) is shown in green, and a setup as the one we use, with $V=31$, in orange. (b) False discovery rate $\operatorname{FDR} = \frac{\textrm{FP}}{\textrm{TP}+\textrm{FP}}$. }
\label{fig:finegrained_fdr_probability}
\end{figure}

\begin{figure*}[t]
\centering
\includegraphics[width=1\linewidth]{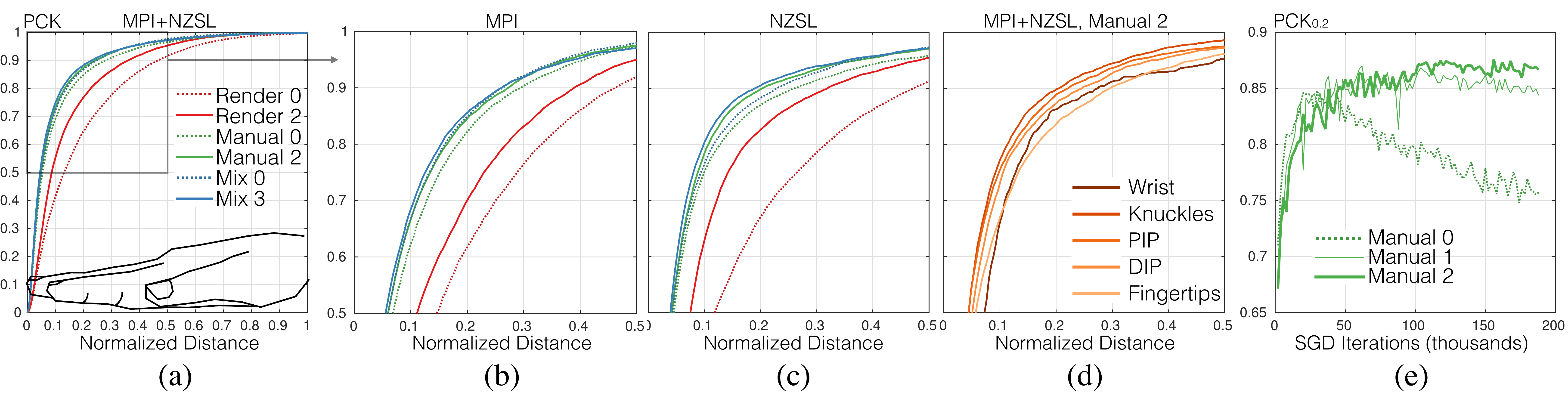}
\caption{Improvement in PCK curves across multiview bootstrapping iterations. (a-c) PCK curves on MPII+NZSL testing images, MPII only, and NZSL only, for two different bootstrapping iterations of each model. Note: improvements are smaller because the evaluation set is from the subset of images that annotators successfully labeled---and the greatest improvements from multiview bootstrapping are usually observed in the complement to this set. (d) PCK for different types of hand joints. (e) Evolution of testing set $\operatorname{PCK}_{0.2}$ with SGD training iterations, for 3 different bootstrapping  iterations. }
\label{fig:pck_iters}
\end{figure*}

\noindent \textbf{Result 1}. The probability of a false triangulation supported by at least $n$ inliers among uniform random 2D points in $V$ views is approximately
$\mathrm{FT}_n\approxeq \operatorname{Pr}(Y \geq 1)$, where $$Y \sim \operatorname{B}\left(\begin{pmatrix}V\\2\end{pmatrix}, q_n\right)$$ is a random variable denoting the number of view pairs supported by at least $n$ inliers. \\

\noindent \textbf{Result 2}. The true and false positive rates for a given keypoint detector $d$ with multiview verification, for a point $p$ across $V$ views with a minimum of $n$ inliers, are approximated by
\begin{align}
\operatorname{TP}_p({d_{}}) &= \operatorname{Pr}(Z \geq n)
\label{eq:hand_tp_single_p}\\
\label{eq:hand_fp_single_p}
\operatorname{FP}_p({d_{}}) &= (1-\operatorname{TP}_p(d_{}))\cdot \mathrm{FT}_{n},
\end{align}
where $Z \sim \operatorname{B}(V,\operatorname{PCK}^p_\sigma(d_{}))$ is a random variable denoting the number of correct 2D detections in $V$ views, and where the incorrect detections are assumed to be uniformly randomly distributed.

\vspace{5pt}
\noindent \textbf{Result 3}. For a complex object with a total of $P$ keypoints, if we require that all keypoints  $p\in[1\dots P]$ be correct to accept a frame and assume that $\textrm{PCK}^p_\sigma({d_{}})$ is the same for all keypoints, then
\begin{align}
\operatorname{TP}({d_{}}) &= \operatorname{TP}_p({d_{}})^P
\label{eq:hand_tp_many_p}
\\\operatorname{FP}({d_{}}) &= \sum_{k=1}^P \begin{pmatrix}P\\k\end{pmatrix} \operatorname{TP}_p({d_{}})^{P-k} \operatorname{FP}_p({d_{}})^k.
\label{eq:hand_fp_many_p}
\vspace{5pt}
\end{align}

Fig.~\ref{fig:finegrained_fdr_probability} shows graphs with varying number of views $V$, minimum number of inliers $n$, and detector quality $\mathrm{PCK}_\sigma$. By generating graphs such as these, recommended number of views can be read out for target false discovery rates or conversely the TP/FP probabilities can be obtained for a given number of views.


\section{Evaluation}
None of the available hand pose estimation datasets we reviewed suited our target use case: general, in-the-wild images containing everyday hand gestures and activities.  We therefore manually annotated two publicly available image sets: (1)~The MPII Human Pose dataset~\cite{Andriluka-14}, which contains images extracted from YouTube videos explicitly collected to reflect every-day human activities, and (2)~Images from the New Zealand Sign Language (NZSL) Exercises of the Victoria University of Wellington~\cite{NZSL}, in which several people use NZSL to tell stories. We chose the latter because it contains a variety of hand poses as might be found in conversation (less common in MPII). Figure~\ref{fig:training_sets} shows a selection of images with manual annotations from both sets. 
To date, we have collected annotations for $1300$ hands on the MPII set and $1500$ on NZSL, which we split $70/30$ into training ($2000$ hands) and testing ($800$) sets. 
\subsection{Improvement with Multiview Bootstrapping}
We evaluate multiview bootstrapping by applying Algorithm~\ref{alg:mvbs} on three initial detectors. All three detectors follow the architecture described in~Sect.~\ref{sect:architecture}, but are trained on 3 different sets of initial training data $\mathcal{T}_0$: (1) ``Render'': an initial set of synthetically  rendered\footnote{We use two renderers, UnrealEngine 4 and a simple raytracer. Characters in UnrealEngine are posed by Mixamo; for the raytracer, we randomly sample hand poses. See the supplementary material for more details.} 
\begin{figure}[t]
\includegraphics[width=\linewidth,clip=true,trim=0 0 450 0pt]{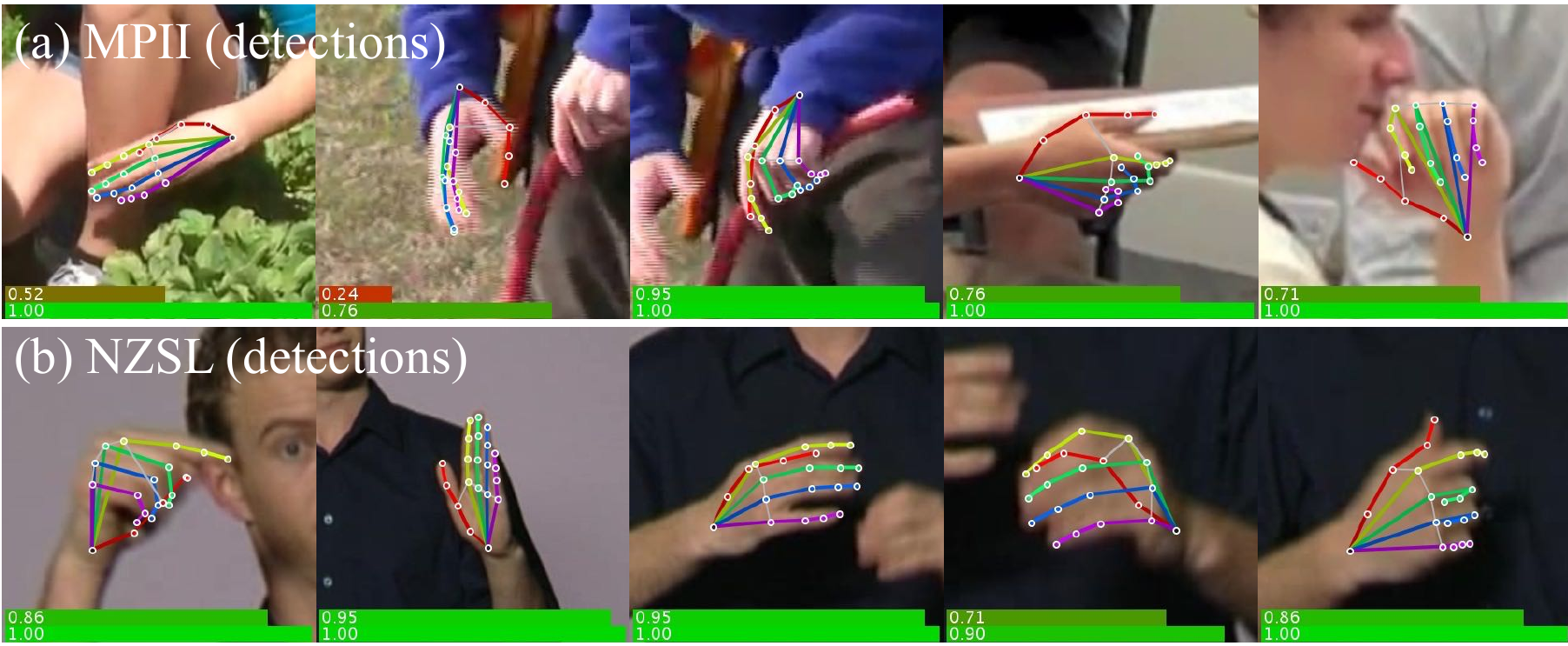}
\caption{Detections using model ``Mix 3'' on test images. To show a representative sample, we pick the first 5 images from the test sets of (a) MPII, and (b) NZSL. Each image's PCK at $\sigma\,{=}\,0.1$ and $\sigma\,{=}\,0.2$ (across the $21$ keypoints) is shown as a pair of bars at the bottom of the image, colored from red (0) to green (1).} 
\label{fig:mpi_nzsl_qualitative}
\end{figure}
images of hands, totalling around $11000$ examples, (2) ``Manual'': the manual annotations in the MPII and NZSL training sets described above, and (3) ``Mix'': the combination of rendered data and manual annotations. For multiview bootstrapping, we use images from the Panoptic Studio dataset~\cite{Joo-15}. In particular, we use 31 HD camera views, and $4$ sequences which contain hand motions, and we use the provided 3D body pose~\cite{Joo-15} to estimate occlusions and bounding boxes for hand detection. When performing bootstrapping iterations, we discard frames with an average number of inliers~$<5$ or an average reprojection error~$>5$, with detection confidence threshold $\lambda{=}0.2$.  
For no iteration did we have to manually discard more than 15 incorrectly labeled frames.

\vspace{3pt}
\noindent \textbf{6.1.1. PCK}. We measure performance as PCK curves averaged over all keypoints, and we evaluate on the testing set of combined MPII and NZSL images ($800$ hands). This is shown in Figure~\ref{fig:pck_iters}a, where we append the bootstrapping iteration to the name of the model, e.g., ``Manual 1'' is the model trained on $\mathcal{T}_0\cup\mathcal{T}_1$, with the initial $\mathcal{T}_0$ built with manual annotations. The PCK curves are plotted by varying the accuracy threshold $\sigma$ in Eq.~\eqref{eq:pck}; this parameter is shown on the horizontal axis. We measure $\sigma$ as a normalized distance, where pixel distances in each example are normalized by $0.7$ times the head size of the corresponding person (approximately the length of an outstretched hand).
Unsurprisingly, the model trained exclusively on rendered data performs worst, but has the most to gain from bootstrapping with real training data. In Fig.~\ref{fig:pck_iters}b and c, we can see that the datasets reflect two levels of difficulty: MPII images vary widely in quality, resolution, and hand appearance, containing many types of occluders, hand-object interactions (e.g., sports, gardening), self-touching (e.g., resting head on hands), as well as hand-hand interactions, as shown in Fig.~\ref{fig:mpi_nzsl_qualitative}a. The NZSL set by contrast is fairly homogeneous, containing the upper-body of people looking directly at the camera and explicitly making visible gestures to communicate (Fig.~\ref{fig:mpi_nzsl_qualitative}b).
Additionally, we study performance for different types of joints in Fig.~\ref{fig:pck_iters}d. These are ordered from closest to the wrist to farthest\footnote{PIP and DIP are the Proximal and Distal Inter-phalangeal joints.}, an order which also corresponds to their  difficulty. Finally, we show how multiview bootstrapping can help prevent overfitting in Fig.~\ref{fig:pck_iters}e, especially for small initial sets $\mathcal{T}_0$.

\begin{table*}[t]
\centering
	\caption{Average 2D error in pixels on the dataset of Tzionas et al.~\cite{Tzionas-16}.}
	\label{Table:quant_dtzmono}
\resizebox{\textwidth}{!}{\begin{tabular}{|l|llll|lllllll|lllllll|}
\hline
 & \multicolumn{4}{|c|}{Single Hand} & \multicolumn{7}{c|}{Hand-Object} & \multicolumn{7}{c|}{Hand-Hand} \\ \hline \hline
Sequence & Grasp  & Flying & \begin{tabular}[c]{@{}l@{}}Rock\\ Gesture\end{tabular} & \begin{tabular}[c]{@{}l@{}}Bunny\\ Gesture\end{tabular} & \begin{tabular}[c]{@{}l@{}}Ball One\\ Hand\end{tabular} & \begin{tabular}[c]{@{}l@{}}Ball Two\\Hands\end{tabular} & \begin{tabular}[c]{@{}l@{}}Bend\\ Pipe\end{tabular} & \begin{tabular}[c]{@{}l@{}}Bend\\ Rope\end{tabular} & \begin{tabular}[c]{@{}l@{}}Ball\\ Occlu.\end{tabular} & \begin{tabular}[c]{@{}l@{}}Move\\ Cube\end{tabular} & \begin{tabular}[c]{@{}l@{}}Moving\\ Occlu.\end{tabular} & Walk   & Cross  & \begin{tabular}[c]{@{}l@{}}Cross\\ Twist\end{tabular} & \begin{tabular}[c]{@{}l@{}}Tip\\ Touch\end{tabular} & Dancing & \begin{tabular}[c]{@{}l@{}}Tip\\ Blend\end{tabular} & Hugging \\ \hline
\cite{Tzionas-16}   & $\mathbf{4.37}$   & $\mathbf{5.11}$   & $4.44$                                                   & $\mathbf{4.50}$                                                     & $6.10$                                                        & $\mathbf{7.15}$                                                        & $6.09$                                                & $5.65$                                                & $\mathbf{8.03}$                                                         & $\mathbf{4.68}$                                                & $5.55$                                                       & $\mathbf{5.99}$   & $\mathbf{4.53}$   & $\mathbf{4.76}$                                                  & $\mathbf{3.65}$                                                & $\mathbf{6.49}$    & $\mathbf{4.87}$                                                & $\mathbf{5.22}$    \\
Ours   & $5.49$ & $5.67$ & $\mathbf{4.15}$                                                 & $4.81$                                                  & $\mathbf{5.75}$                                                     & $9.79$                                                      & $\mathbf{5.47}$                                              & $\mathbf{4.35}$                                               & $9.66$                                                       & $6.38$                                              & $\mathbf{5.40}$                                                     & $9.10$ & $6.95$ & $10.09$                                               & $5.31$                                              & $6.55$  & $6.09$                                              & $10.35$ \\ \hline
\end{tabular}}
\end{table*}

\vspace{3pt}
\noindent \textbf{6.1.2 Robustness to View Angle}.
\begin{figure}
\includegraphics[width=\linewidth]{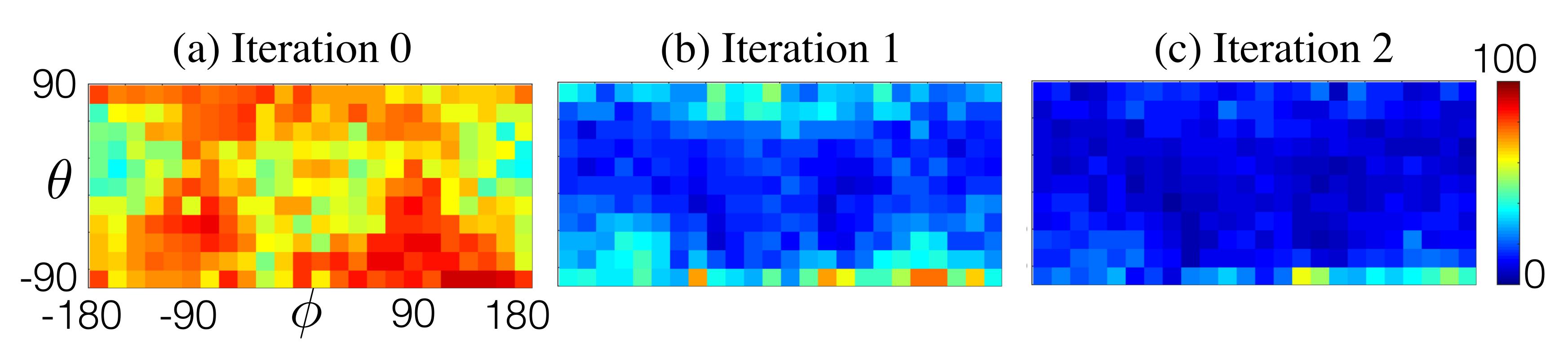}
	\caption{Robustness to view angle. We show percentage of outliers as a heatmap for each viewing angle, where azimuth $\phi$ is along the X axis and elevation $\theta$ along the Y axis.}\label{fig:viewRobustness}
	\vspace{-12pt}
\end{figure}
We quantify improvement in the detector's robustness to different viewing angles by measuring the percentage of outliers during 3D reconstruction. As ground truth, we visually inspect our best 3D reconstruction result and select only correctly reconstructed frames.  We define view angle as azimuth $\phi$ and elevation $\theta$ w.r.t a fixed hand at the origin. Intuitively, angles with $\phi{=}\{-180,0,180\}$ (viewing the palm or backhand face-on) are easier because there is less self-occlusion. At $\phi{=}\{-90,90\}$, we are viewing the hand from the side, from thumb to little-finger or vice-versa, resulting in more occlusion. Similarly, at $\theta{=}\{90,-90\}$ the viewing angle is from fingertips to wrist, and vice-versa; these are the most difficult viewpoints. We compare the first iterations of the ``Mix'' detector, which quickly becomes robust to view diversity. We plot this as a heatmap, where we bin hand detections using each example's azimuth and elevation. The percentage of outliers is computed using all the examples that fall in each bin.

\subsection{Comparison to Depth-based Methods}

We quantify the performance of our method on a publicly available dataset from Tzionas et al.~\cite{Tzionas-16}. Although there exist several datasets often used to evaluate depth-based methods, many of them do not have corresponding RGB images, or their annotations are only valid for depth images. Datasets with RGB images and manual annotations that can be accurately localized are rare; the dataset from~\cite{Tzionas-16} is the best match to quantify our method\footnote{Some other datasets also have manual keypoint annotations with RGB images~\cite{Sridhar-16, Tompson2014b}, but the calibration parameters in~\cite{Sridhar-16} were not accurate enough, and the images in~\cite{Tompson2014b} are warped to match depth.}. We run the 2D keypoint detector ``Mix 3'' on the RGB images of the dataset. The sequences include single-hand motion, hand-hand interaction, and hand-object interaction. To allow direct comparison with~\cite{Tzionas-16}, we use average pixel errors in the location of the keypoints provided, as shown in Table~\ref{Table:quant_dtzmono}. Note that the method of~\cite{Tzionas-16} is based on a complex 3D hand template and uses depth data and tracking, taking several seconds per frame. Our result shows comparable performance for single-hand and hand-object scenarios, using only per-frame detection on RGB and can be run in realtime using GPUs. Performance degrades for hand-hand interaction: When one of the hands is very occluded, our detector tends to fire on the occluding hand. Detecting joints on both hands simultaneously would be advantageous in these cases, rather than treating each hand independently as our current approach does.
\begin{figure}[t]
\centering
\includegraphics[width=\linewidth,clip=true,trim=0 5pt 0pt 5pt]{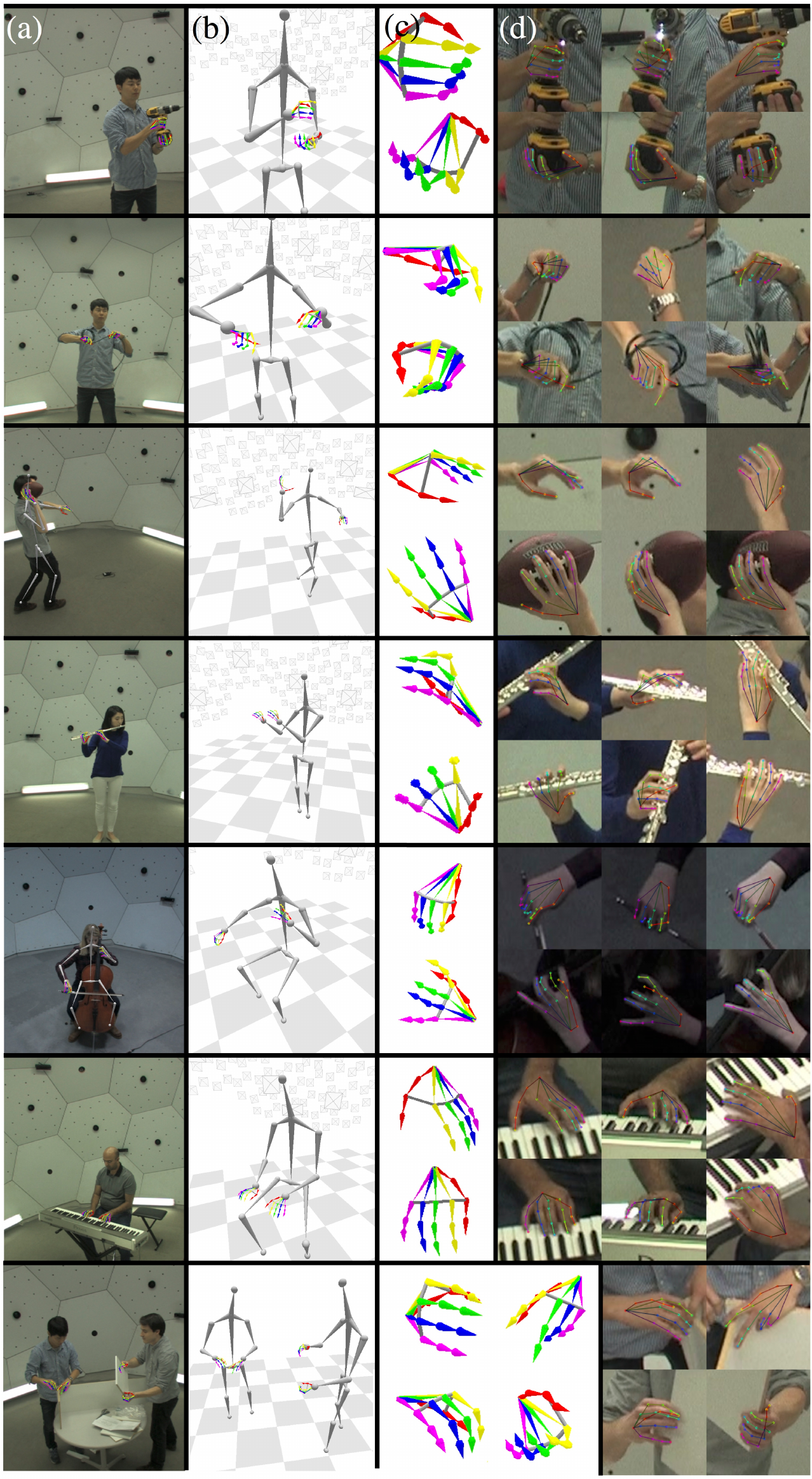}
\caption{Qualitative multiview results on sequences not used during bootstrapping. (a)~Reprojected triangulation. (b)~3D hands in context. (c)~Metric reconstruction. (d)~2D~detections from ``Mix 3'' on selected views. }
\label{fig:qual3d_singles_comp}
\end{figure}
\begin{figure}[t]
\centering
\includegraphics[width=\linewidth]{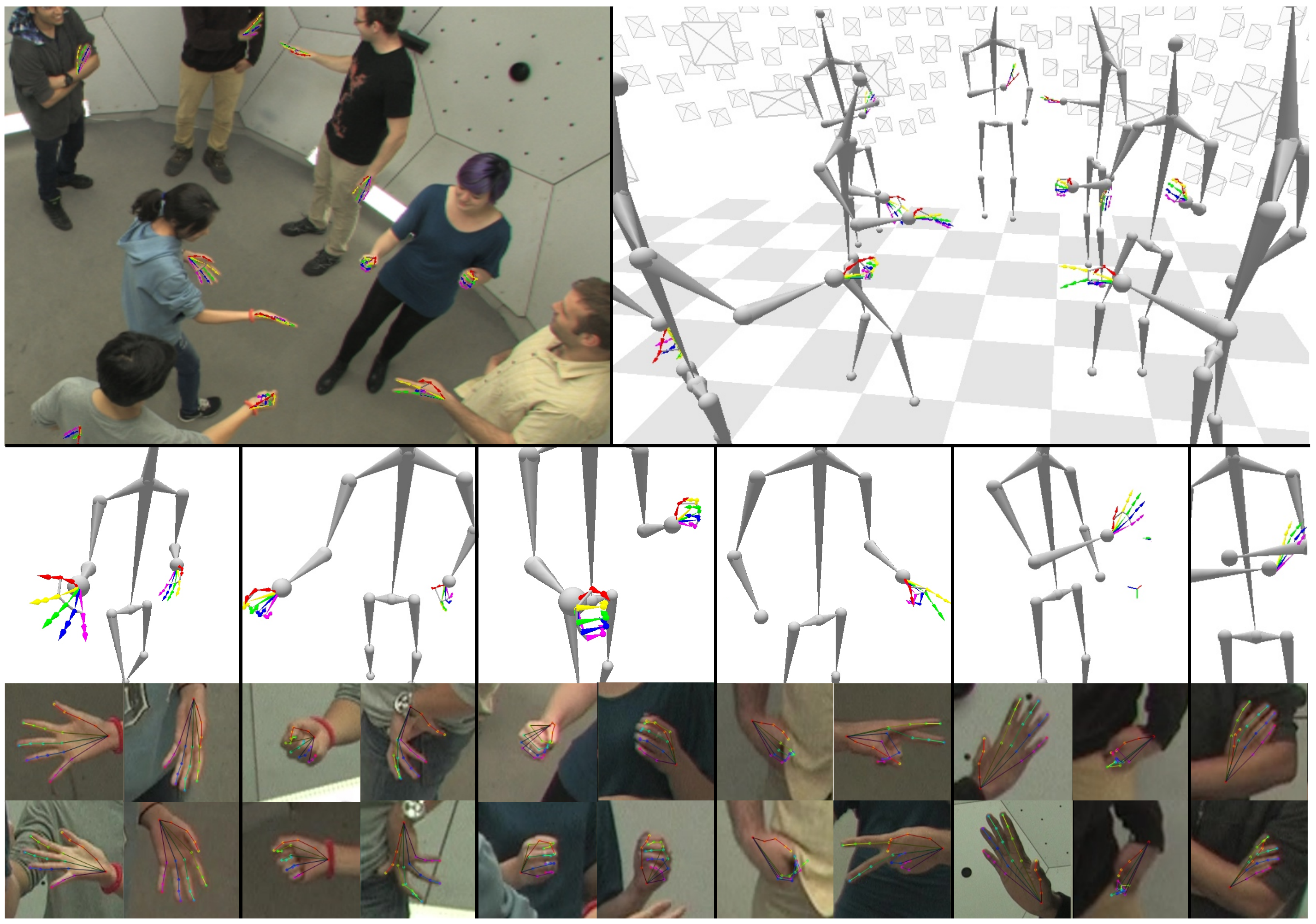}
\caption{A keypoint detector that works at typical camera resolutions combined with a multiview system allows capturing the hand motions of entire groups of interacting people, something that was not possible with prior approaches.}
\label{fig:qual3d_goup_comp}
\end{figure}

\subsection{Markerless Hand Motion Capture}
The trained keypoint detectors allow us to reconstruct 3D hand motions in various challenging scenarios. We use the ``Mix 3''  detector on 31 HD camera views on Panoptic Studio data~\cite{Joo-15}, and generate 3D hands by triangulation as we do for multiview bootstrapping. Test scenes include various practical hand motions, such as manipulating diverse tools (e.g., drill, scissors, ruler), sports motions (throwing a ball, bat swing), and playing musical instruments (piano, cello, flute, and guitars). We also reconstruct scenes with multiple interacting people including social games, shelf building, and a band performance. Some 3D hand reconstruction results with corresponding 2D detections are shown in Figures~\ref{fig:qual3d_singles_comp} and~\ref{fig:qual3d_goup_comp}. Note that most depth-based methods are not applicable in these scenarios due to short sensor ranges and difficulties handling hand-object interactions. The results are best viewed in the supplementary video, which includes additional reconstruction results. 
\vspace{-4pt}
\section{Discussion}
This paper presents two innovations: (1) the first realtime hand keypoint detector showing practical applicability to in-the-wild RGB videos; (2) the first markerless 3D hand motion capture system capable of reconstructing challenging hand-object interactions and musical performances without manual intervention. We find that rich training sets can be built using multiview bootstrapping, improving both the quality and quantity of the annotations. Our method can be used to generate annotations for any keypoint detector that is prone to occlusions (e.g., body and face). Building a large annotated dataset is often the major bottleneck for many machine learning and computer vision problems, and our approach is one way to refine weakly supervised learning by using multiview geometry as an external source of supervision. As future work, making the method robust enough to work with fewer cameras and in less controlled environments (e.g., with multiple cellphones) would allow the creation of even richer datasets that more closely reflect real world capture conditions.

{\small
\bibliographystyle{ieee}
\bibliography{main}
}

\end{document}